\documentclass{article}

\usepackage{PRIMEarxiv}

\usepackage[utf8]{inputenc} 
\usepackage[T1]{fontenc}    
\usepackage{hyperref}       
\usepackage{url}            
\usepackage{booktabs}       
\usepackage{amsfonts}       
\usepackage{nicefrac}       
\usepackage{microtype}      
\usepackage{lipsum}
\usepackage{fancyhdr}       
\usepackage{graphicx}       
\graphicspath{{media/}}     
\usepackage[linesnumbered,ruled,vlined]{algorithm2e}
\usepackage{amsmath}

\title{Subgroup Discovery in Unstructured Data
}

\author{
  Ali Arab \\
  Simon Fraser University \\
  Canada\\
  \texttt{aarab@sfu.ca} \\
   \And
  Dev Arora \\
  Simon Fraser University \\
  Canada\\
  \texttt{dev\_arora@sfu.ca} \\
   \AND
   Jialin Lu \\
   Simon Fraser University \\
   Canada \\
   \texttt{jialin\_lu@sfu.ca} \\
   \And
   Martin Ester \\
   Simon Fraser University \\
   Canada \\
   \texttt{ester@sfu.ca} \\
}

\begin{document}
\maketitle

\begin{abstract}
Subgroup discovery is a descriptive and exploratory data mining technique to identify subgroups in a population that exhibit interesting behavior with respect to a variable of interest. Subgroup discovery has numerous applications in knowledge discovery and hypothesis generation, yet it remains inapplicable for unstructured, high-dimensional data such as images. This is because subgroup discovery algorithms rely on defining descriptive rules based on (attribute, value) pairs, however, in unstructured data, an attribute is not well defined. Even in cases where the notion of attribute intuitively exists in the data, such as a pixel in an image, due to the high dimensionality of the data, these attributes are not informative enough to be used in a rule. In this paper, we introduce the subgroup-aware variational autoencoder, a novel variational autoencoder
that learns a representation of unstructured data which leads to subgroups with higher quality. Our experimental results demonstrate the effectiveness of the method at learning subgroups with high quality while supporting the interpretability of the concepts.
\end{abstract}

\keywords{Subgroup Discovery \and Unstructured Data \and  Pattern Mining}

\section{Introduction}

Subgroup discovery is a descriptive data mining technique aimed at identifying descriptions of a subset of the data that show interesting behaviors with respect to a variable of interest, the target variable, through behaviour deviates from the behaviour of the population as a whole \cite{novak2009supervised, lavrac2004subgroup, klosgen1996explora, herrera2011overview, helal2016subgroup}. Subgroup discovery methods are used for data exploration, automatic hypothesis generation, and more importantly, for extracting actionable knowledge for decision support. A prominent example is knowledge discovery in medical domains, where insights on treatment response for different subgroups can facilitate personalized treatment plans \cite{nannings2008subgroup}. For instance, a subgroup in this context may be stated as the following expression:

\begin{equation}
     \textit{geneder=female} \; \; AND \; \; \textit{age}<30 
\end{equation}


which would be considered interesting if the success rate in this subgroup is significantly lower or higher than the success rate in the dataset as whole.  

Despite numerous works in the literature, all subgroup discovery techniques suffer from a limitation - they are only applicable to structured data, i.e., data with well-defined attributes. A typical instance of this would be data found relational databases. However, in many settings, the data to be dealt with is unstructured and potentially high dimensional such as images, videos, text documents, and time series. Subgroups are defined through a conjunction of attribute and value pairs. For high dimensional data, such as an image, this implies that the subgroups will be defined based on the value of individual pixels. This approach will fail because the value of individual pixels is not informative enough to be used in a subgroup. Instead, the collective behaviors of pixels need to be evaluated to extract interesting subgroups.

Although grouping pixels might appear like a plausible approach to address the above-mentioned issue, this method still performs poorly if an important aspect of the data is not localized, but instead is  spread irregularly over many input features. For example, in a portrait image, the characteristic "age" affects the whole area of the face, as opposed to one specific area. As another example, in time series, properties such as trend and seasonality are non-local features. In such scenarios, the attribute and value pairs need to be defined using concepts that capture high-level information in the data. For example, a concept could refer to a time series being periodic, or a time series having high frequencies. In the case of images, a concept could refer to the shape or color of an object.



Variational autoencoders (VAEs) have shown promising results in learning disentangled representations of the factors involved in the data generation process that could serve for defining concepts\cite{higgins2016beta, hou2017deep}. In this paper, we propose a VAE with a customized loss function to approximate the true generation process; subgroups are then defined in this latent space. The customized loss function encourages the VAE to learn a subgroup-friendly representation of the data.



Ideally, a subgroup discovery loss value needs to be added to the VAE loss function tp makes the optimization framework aware of the operations performed  on the input to compute the subgroups. However, subgroup discovery involves several non-differentiable operations that prevent us from using it directly in the loss function. As an alternative, we propose using a surrogate loss based on the correlation of the latent variables with the target variable which is easy to compute. In the forward pass, a latent representation is computed for a batch of training examples, then subgroup discovery is applied to the latent representation to identify the top subgroups. Finally, the correlation of the target variable and the latent variables present in the top subgroups is computed as the surrogate subgroup discovery loss. In the backward pass, a sum of the reconstruction loss and the computed surrogate subgroup discovery loss is used to update the weights of the VAE. The reconstruction loss updates the weights for both encoder and decoder, while the subgroup discovery loss affects only the encoder of the network.

Our main contributions are the following:

\begin{itemize}
    \item We propose a general framework to discover subgroups in unstructured high-dimensional data. 
    \item We present a subgroup-aware VAE that learns a representation of the data that leads to subgroups with high quality.
    \item We demonstrate the effectiveness of our approach in identifying subgroups of a real-world dataset with quantitative and qualitative analysis. 
\end{itemize}


\section{Background and Related Works}

Discovery of actionable knowledge from unstructured data remains an essential task for a number of critical domains \cite{giudice2019approach, abdualgalil2020efficient}. This paper focuses on discovering subgroups from unstructured data - discovering high-level concepts (features) and utilizing such concepts to identify subgroups with statistical significance w.r.t certain variables of interest. Therefore, our work is closely related to the topics of concept discovery and subgroup discovery. In the following two subsections, we review these two topics.

\subsection{Subgroup Discovery}
Given a population of individuals and a property of interest, subgroup discovery aims to identify subgroups that are statistically "most interesting". The interestingness measure depends on the application, but conventionally is designed to capture subgroups with the most unusual distribution with respect to the property of interest.

We define formally the task of subgroup discovery with a quadruple \\ $(D, \sum, T,Q)$. $\mathcal{D}$ specifies the dataset, $\sum$ is the search space of candidate subgroups, $T$ is the target variable of interest, and $Q$ is the interestingness function, a function that measures the quality of the subgroups.

A dataset $D=(\mathcal{I},\mathcal{A})$ is given by a set of individuals $\mathcal{I}=c_{1}, c_{2}, \ldots, c_{n}$, and a set of attributes $\mathcal{A}=A_{1}, A_{2}, \ldots, A_{m}$, where $n$ is the number of individuals, and $m$ is the number attributes. Encoding characteristics associated with individuals, each attribute $A_j: \mathcal{I}\rightarrow \operatorname{range}\left(A_{j}\right)$ is a function that maps each individual to a a value in its range. Value of attribute $A_j$ for individual $c_i$ can be denoted as $A_{j}(c_i)$.

The search space $\Sigma$ is a set of subgroup descriptions, where each subgroup description is formed based on selectors. A selector sel: $\mathcal{I} \rightarrow\{$ true, false $\}$ is a boolean function that filter individuals based on their attribute values. As an example, the selector age $=$ 30 returns a boolean vector whose values are true for all individuals with age 30.

For easier interpretation, a subgroup description, denoted by $P$, is most commonly defined as a conjunctive combination of selectors, i.e., $P=\operatorname{sel}_{1} \wedge \operatorname{sel}_{2} \wedge \ldots \wedge \operatorname{sel} _{d}$. The set of all individuals for which this expression evaluates to true is called the subgroup cover. The set $sg(P)=\{c \in \mathcal{I} \mid P(c)=$ true $\}$ denotes all individuals which are covered by P, i. e., individuals for which the conjunctive expression is evaluated to true. The empty subgroup description $P_{\emptyset}=\emptyset$ covers all instances in the dataset. The count of individuals in each subgroup is denoted by $i_{P}=|sg(P)|$.

The interestingness measure $Q: P \rightarrow  \mathbb{R}$  is a function that maps every subgroup to a real number which indicates the quality or interestingness of the subgroup. Subgroup discovery algorithm return $K$ subgroups which have the highest value in terms of interestingness. In general, interestingness functions compare statistical distribution of the target variable $T$ with that of the subgroup to compute a score. Typically there is a trade-off between size of a subgroup and its deviation $\tau_P-\tau_\emptyset$, where $\tau_P$ is the average value of $T$ in the subgroup $P$ and $\tau_\emptyset$ is the average value of $T$ in te whole population. The lower the pursued size, the higher the chance of obtaining more interesting subgroups, however, it is difficult to claim a general rule if the size of the subgroup is low. In the extreme case, the algorithm may return a subgroup consisting of only one sample with a very high score, but there is no value in this subgroup. Therefore, the interestingness measure is traditionally defined based on the following formula:

\begin{equation} \label{eq:q_func}
    Q(P)  = (\frac{i_P}{N})^\alpha (\tau _P - \tau _\emptyset)
\end{equation}

where $i_p$ is the subgroup size, $N$ is the number of individuals in the dataset, and $\alpha$ is the parameter adjusting the importance of coverage. Low values for $\alpha$ result in subgroups with high deviation in the target share, even if few samples are covered by the subgroup.  High values of $\alpha$ lead to subgroups with a large sample size, although the deviation of the target share may not be significant \cite{klosgen1996explora}.

For binary target variables, $\tau_P$ can be written as $\tau=\frac{tp(P)}{i_P}$ where $tp(P)$ is the number of individuals in the subgroup for which $T$ is one. In this binary setting, $\tau_P$ is referred as subgroup \textit{target share}.

%
Having defined all four components of a subgroup discovery task, a subgroup discovery algorithm is utilized to search for subgroups with the highest quality. Although an exhaustive search based algorithm guarantees finding the optimal solution, these are not feasible when the search space is large. For time-sensitive applications, a heuristic search such as beam search based and genetic based search algorithms are utilized \cite{clark1989cn2, del2007evolutionary}.

In particular, beam search based methods start with a list of subgroup hypotheses of size $\beta$, known as the beam width, which initially contains an empty subgroup $P_\emptyset$. The subgroups in the beam are then expanded iteratively, and only the top $\beta$ subgroups are kept for the next iteration. The expansion of each subgroup is done by adding a selector to the subgroup. The other parameter is maximum depth which limits the number of selectors in the subgroups. The full algorithm is presented in Algorithm \ref{alg:search}.


\begin{algorithm}[h]
    \KwIn{ F: Features , depth: maximum depth, $\beta$: Beam Width }
    \KwOut{  $output\_beam$, a set of subgroups}
$\textit{selectors}=\textit{create\_selectors }(\textit{F})$ \; 
$\textit{n}=\textit{length }(\textit{selectors})$ \; 
$ \textit{beam} = \textit{heap\_max }(\textit{size}=\beta)$ \;
$\textit{expanded} = \textit{map()}$\;
\For {$d: 1 \textit{ to depth }- 1$}{
    $\textit{output\_beam} = \textit{beam.copy()}$\;
    \For{$C  \textit{ in beam }$}{
        $if ( \textit{expanded}[C]==\textit{True}) \textit{ continue}$\;
        \For{$j: 1 \textit{ to } n$}{
            $I=\textit{selectors[j]}$\;
            $\textit{score} = Q(beam_{i}\cup I) $\;
            $\textit{new\_beam.add }(beam_{i}\cup j),\textit{score})) $\;
        }
        $\textit{expanded}[C]= \textit{True}$\;
    }
    $\textit{beam} = \textit{new\_beam.copy()}$\;
}
\Return  $\textit{beam}$\;
\caption{Beam-search based subgroup discovery algorithm}
\label{alg:search}
\end{algorithm}

A crucial step in subgroup discovery is the definition of meaningful and effective selectors. For nominal attributes, each selector can be defined based on values in the domain of the attribute: $\operatorname{sel}_{A_{j}=v}(c)=$ true $\Leftrightarrow$ $A_{j}(c)=v .$ For example, for the attribute gender, a possible selector is $\textit{sel}_{\textit{gender=male}}$. For numerical attributes, selectors are defined based on an interval in the attribute domain: $\operatorname{sel}_{ A_{j} \in [lb, ub]}(c)=$ true $\Leftrightarrow lb<A_{j}(c) \wedge A_{j}(c)<ub .$, where $lb$ and $ub$ are lower bound and upper bound for the interval, respectively.

Despite extensive works in the subgroup discovery literature, they all suffer from a significant limitation: current approaches to define selectros are limited to structured data. Our goal is to define selectors that can capture high-level semantic patterns in unstructured, high-dimensional data such as images and text documents \cite{herrera2011overview, helal2016subgroup, novak2009supervised}.

\subsection{Concept Discovery from unstructured data}

Concept discovery is a subfield of Knowledge Discovery in Databases (KDD) research that focuses on discovering insight into the underlying conceptual structure of the data \cite{poelmans2010concept}.
The intuition is that such a structure can aid human users in understanding the data and solving various computation tasks and is especially important in critical domains \cite{scharpf2019towards, sacarea2018formal, zatsman2021new}.

Discovering high-level concepts/features to describe unstructured data thus remains a highly needed and technically challenging task.
This is an aspect that modern machine learning techniques often overlook, as modern machine learning approaches, typically deep learning techniques, choose to bypass the problem of concept discovery, as layers of stacked computation plus back-propagation are utilized to enable machines to learn without careful manual feature engineering \cite{lewis1992feature, zheng2018feature}.

Such a shortcut results in highly efficient training and accurate prediction performance on high dimensional unstructured data. However, the ability to understand the data and consequently the ability to explain and control the behavior of the machine are sacrificed.
An improved prediction ability does not necessarily help us better understand the data, as we do not understand the concepts and key information that contribute to the prediction \cite{weld2019challenge}. Thus, an entire subfield of machine learning, interpretability and explainability, are dedicated to address this issue \cite{carvalho2019machine}. 

Often these high-level concepts are obtained through post-hoc approaches, i.e., applied on an already-trained Deep Neural Network. These works start with the assumption that the various human-understandable concepts are somehow encoded as feature vectors that are separable in the subspace of the hidden layers of the model. And that the similarity measures in the subspace become more aligned with human-understandable high-level concepts \cite{cohen2020separability}.
\cite{kim2018interpretability} use a handful of extra human-annotated images to learn a linear discriminator to predict the presence of certain concepts. 
\cite{zhou2018interpreting} measures the possible correlation between concept annotation and neuronal activation to find neurons that are dedicated to detecting certain concepts. 
\cite{radford2017learning} takes a similar approach to generative models. Once the model is trained, exhaustive search and testing are conducted to test the correlation between neuronal activation and the concept annotation of interest. This approach is able to find a specific neuron that indicates the sentiment of the generated sentences.
These post-hoc approaches obtain representations to help better understand the prediction made by the black box neural network \cite{zhou2018interpretable, singh2018hierarchical}. 
It is worth noting that these works typically need to use extra annotations of concepts, though certain works manage to overcome that by utilizing tricks, such as clustering to automatically discover concepts \cite{singh2018hierarchical,ghorbani2019towards}.

Alternative to post-hoc approaches, many works attempt to discover possibly human-aligned latent variables by specifying the right regularization \cite{higgins2016beta} or architectural constraints \cite{singh2020transformation, rhodes2021local}.
However, \cite{locatello2019challenging} conducted a massive scale experiment and argue the impossibility to obtain meaningful latent features in a completely unsupervised setting. In our work, we use the variable of interest commonly used in subgroup discovery as the extra label to provide indirect supervision.

Aside from discovering concepts from unstructured data, existing works are often applied and analyzed on the basis of a single data instance and neglect the information shared among different data instances. Subgroup discovery, as covered in the above subsection, is an exemplar field where the need to mine unstructured data to find interesting feature descriptors and groups of data instances w.r.t certain variables of interest remains ubiquitous. And yet recent approaches do not consider the problem in its workflow and traditional approaches are simply inapplicable to unstructured data.





\section{The Proposed Method}

In this section, we propose the subgroup-aware VAE.  We first revisit the basic VAE, which forms the basis of our method. Then the subgroup-aware VAE is introduced, which learns a “subgroup-friendly” latent representation by adding a corresponding term to the loss function. Finally, a subgroup discovery algorithm is applied to detect the most interesting subgroups based on the learned latents.

\subsection{Basic VAE}
Autoencoder is an unsupervised learning technique that leverages neural networks to learn a compact representation of the data. Autoencoder consists of an encoder component, which transforms the data to a low dimensional vector, known as a latent vector, and a decoder component which builds the original input from the latent vector. The network can be trained by minimizing the reconstruction error, $\mathcal{L}(x, \hat{x})$, which measures the difference between the original input $x$ and the reconstruction $\hat{x}$. 

As a modern variant, variational autoencoders interpret the latent nodes as probabilistic distributions corresponding to the factors in the data generation process. The loss function for VAEs, as shown in equation \ref{eq:vae_loss}, has two components. In addition to the reconstruction loss, there is another term added: the  Kullback–Leibler (KL) divergence between learned distribution $q$ and the prior distribution $p$, for each latent dimension $j$. This additional term regularizes the latent distributions to be close to the prior distribution. The prior distribution is typically selected to be an isotropic unit Gaussian $\mathcal{N}(0,1)$.

\begin{equation}
\label{eq:vae_loss}
\mathcal{L}(x, \hat{x})+\sum_{j} K L\left(q_{j}(z \mid x) \| p(z)\right)
\end{equation}

\subsection{Subgroup-aware VAE}
To learn a representation that yields interesting subgroups, we need selectors that are a predictor of the target variable. To this end, we add a third component to the VAE loss function that captures the correlation of a subset of latents with the target variable. In the extreme case, one might capture the correlation loss based on all latents, however, this will lead to a poor reconstruction. Therefore, a subset of latents needs to be strategically chosen to improve the performance of subsequent subgroup discovery, while keeping the reconstruction error low. To achieve this goal, in each iteration of training, we perform subgroup discovery in the latent space with respect to the target variable and find the top $k$ subgroups. Then, the latents present in the top $k$ subgroups will be used to compute the correlation values. These correlation values are basis for defining our subgroup discovery loss. The total loss function is then defined as the following: 

\begin{equation}
\label{eq:sa_vae_loss}
\mathcal{L}(x, \hat{x})+\sum_{j \in L} K L\left(q_{j}(z \mid x) \| p(z)\right) + \lambda* \frac{1}{len(L_{SD})}\sum_{j \in L_{SD}} S\left( z_j, T \right)
\end{equation}

where $L$ indicates the set of all latents, $L_{SD}$ refers to the set of latents present in the top $k$ subgroups, $S\left( z_j, T \right)=1-\textit{correlation}(z_j,T)^2$ is a correlation-based loss function for latent $z_j$, and $\lambda$ is a hyperparameter regulating the importance of subgroup discovery loss. We refer to  $\frac{1}{len(L_{SD})}\sum_{j \in L_{SD}} S\left( z_j, T \right)$ as subgroup discovery (SD) loss during the rest of this paper.  

To perform subgroup discovery in the latent space, the continuous latent values  are discretized. Given that the distribution of latent values per latent in the VAE is Gaussian, for a dimension j, we use the following binning rule:

\begin{equation}
  f(l)=\begin{cases} 
      -1 & l\leq mean(z_j)-std(z_j) \\
      0 & mean(z_j)-std(z_j)\leq l\leq mean(z_j)+std(z_j) \\
      1 & mean(z_j)+std(z_j)\leq l 
   \end{cases}
\end{equation}

This binning method allows us to clearly define intuitive areas in each latents range when constructing and evaluating subgroups. The full training algorithm is presented in Algorithm \ref{alg:train} and \ref{alg:sdloss}.

\subsection{Subgroup Discovery}
Once the model is trained using Algorithm \ref{alg:train}, a beam search-based subgroup discovery method, as described in Algorithm \ref{alg:search}, guided by interestingness measure defined in Equation (\ref{eq:q_func}), is applied to the latent representation of all samples to identify the subgroups. The resulting subgroups are subpopulations that satisfy a concept-based constraint in the form of $\wedge^d_1 C_i==v$, where $C_i$ denotes the $i^th$ concept, $v\in \{-1,0,1\}$ is the value of the concept $i$ , and $d$ is the number of selectors in a subgroup descriptor.

\begin{algorithm}
\DontPrintSemicolon
\KwIn{X: Training Features, T:Target variable, epochs: number of epochs\; k: Number of top subgroups to optimize; $\lambda$: Coefficient of SD loss}
\KwOut{sgs: set of subgroups} 
$\textit{VAE}=\textit{init\_model}()$ \; 
\For{$e: 1 \textit{ to epochs }$}{
    \For{$\textit{x, y} \; \textit{in} \; 
    \textit{minibath(A,T)}$}{
        $\textit{recon\_x}=\textit{VAE}(x)$ \;
        $\textit{recon\_loss}=\textit{MSE\_Loss}(x,recon\_x)$ \;
        $\textit{Z}=\textit{VAE.encoder}(x)$ \;
        $\textit{batch\_subgroups}=\textit{find\_subgroups}(Z, y)$ \;
        $\textit{latents\_to\_optimize}=\textit{get\_latents\_in\_subgroups}(batch\_subgroups)$ \;
        $\textit{SD\_loss}=\textit{compute\_SD\_loss}(Z, latents\_to\_optimize, y) \;\#\textit{Algorithm 3}$ \; 
        $\textit{total\_loss}=\textit{recon\_loss}+ \lambda * \textit{SD\_loss}$ \;
        $\textit{total\_loss.computeGradients}()$ \;
        $\textit{VAE.updateWeights}()$ \;
    }
}

\caption{Training algorithm}
\label{alg:train}
\end{algorithm}


\begin{algorithm}
\DontPrintSemicolon
\KwIn{Z: Batch latent vector, latent\_ids: set of latents to optimize, T: Target}
\KwOut{loss: SD loss value} 

$sum\_loss = 0$ \;
\For{$latent\_id  \; \textit{in} \; \textit{latent\_ids}$}{
        $corr=\textit{compute\_correlation}(Z[latent\_id], T)$ \; 
        $loss=1- corr^2$ \;
        $sum\_loss=sum\_loss + loss $ \;
}
$\textit{SD\_loss}=\frac{sum\_loss}{len(latent\_ids)}$ \;
\Return  $\textit{SD\_loss}$\;
\caption{Subgroup discovery loss function}
\label{alg:sdloss}
\end{algorithm}






\section{Experimental Evaluation}
In our experimental evaluation, we answer the following research questions:
\begin{itemize}
    \item How does the quality of the subgroups extracted from the subgroup-aware VAE differ from those extracted from a basic VAE in terms of coverage and target share?
    \item Are the extracted subgroups from the subgroup-aware VAE human-understandable?
\end{itemize}

To answer these questions, we performed experiments with three different VAE models:
\begin{itemize}
    \item \textbf{VAE}: training the model with only the original VAE loss.
    \item \textbf{VAE+SD F.T.}: pre-training the model with the original VAE loss, and then fine-tuning the model with subgroup-discovery loss. 
    \item \textbf{VAE+SD}: training the model from scratch with subgroup discovery loss and the original VAE loss.
\end{itemize}

\subsection{Dataset}

The dataset we used for running the experiment is CelebA, which is a large-scale face dataset with 202,599 face images, and 40 binary attributes per image \cite{liu2015faceattributes}. As a preprocessing step, images were resized to 64x64 pixels. The attributes attractiveness and gender (males encoded as one) were used as the target variable.

\begin{figure*}[h!]
\centering
\includegraphics[scale=0.25]{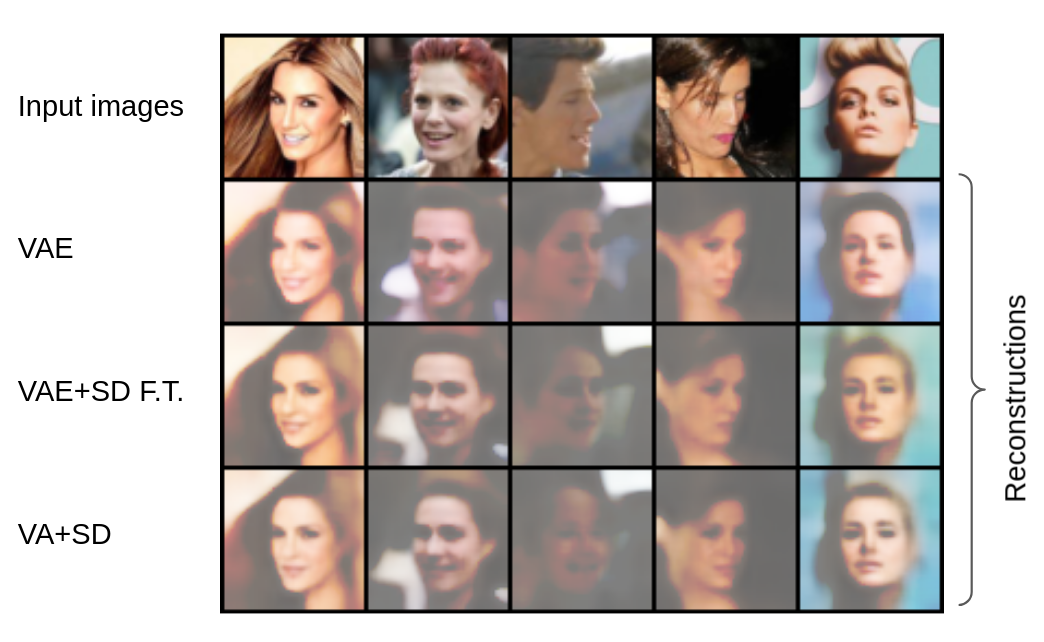}
\caption{Reconstructed images for the three models}\label{fig:reconstructions}
\end{figure*}

\subsection{Hyperparameters and implementation}
The Adam optimizer was used to train the model. The hyperparameters used in the experiment are as follows: beam-width $\beta=10$, coverage importance coefficient $\alpha=0.5$, number of top subgroups to optimize $k=1$, SD loss coefficient $\lambda=10$, and learning rate = 0.001. 

The different VAE models were implemented in Python using pyTorch library. The source code is publicly available at \url{https://github.com/subgroupawarevae/savae}. We used  NVidia A100 GPUs to run the experiments.





\subsection{Quality of the Subgroups}\label{sgd_quality_section}
To evaluate the quality of the subgroups, we computed the subgroup quality measures coverage and target share, which are standard performance metrics for subgroup discovery. Table \ref{tab:sgd_quality} shows the quality measures for the baseline basic VAE, VAE+SD F.T., and VA+SD. The ratio of positive labels in the attractiveness target variable 0.51, and the proportion of males is 0.49, therefore subgroups with target share values further away from these values are considered more interesting. We can see that in terms of target share, both models that use subgroup discovery loss outperform the baseline while training the model from scratch with SD loss yields the highest improvement. In terms of coverage, we also see a slight improvement in comparison to the baseline, with the model with fine-tuning having the highest value.
The other observation is that the improvement in target share for the gender target variable is significantly higher in comparison to the gain in attractiveness. It could be due to the fact that attractiveness is a very subjective attribute and there are no hard rules to define that. On the other hand, characteristics to distinguish males vs. females are easier to identify, so the model is able to find the relevant subgroups more easily.

As expected, adding the subgroup discovery loss increases the reconstruction error. To better understand the effect of SD loss on reconstruction quality, we visualized the reconstruction of 5 sample images for each model. As shown in Figure \ref{fig:reconstructions}, we can see that the reconstructions are still acceptable even for the model that is trained with SD loss from scratch. However, looking carefully at the images, we may observe that some characteristics of the face in the original image have been lost in VAE+SD; for example in the first image, the smile of the person has not been reconstructed properly. 

Considering both the gain in coverage and target share and the increase in reconstruction error, the model with SD loss and fine-tuning seems to provide the best trade-off. 

\begin{table}[]
\begin{tabular}{l|ccc|ccc}
                                & \multicolumn{3}{c}{Attractiveness}             & \multicolumn{3}{c}{Gender}                     \\
Method                          & Target Share & Coverage & Recon. Error & Target Share & Coverage & Recon. Error \\
\specialrule{.2em}{0em}{.2em}
VAE                             & 0.771        & 0.023    & 7593.6               & 0.784        & 0.025    & 7591.8               \\
VAE+SD F.T.          & 0.833        & 0.028    & 7613.4               & 0.990        & 0.038    & 7614.8               \\
VA+SD & 0.953        & 0.027    & 7708.7               & 0.993        & 0.037    & 7619.3              
\end{tabular}
\caption{Subgroup quality measures and reconstruction error for the three model configurations} \label{tab:sgd_quality}
\end{table}

\begin{figure*}[h!]
\centering
\includegraphics[scale=0.5]{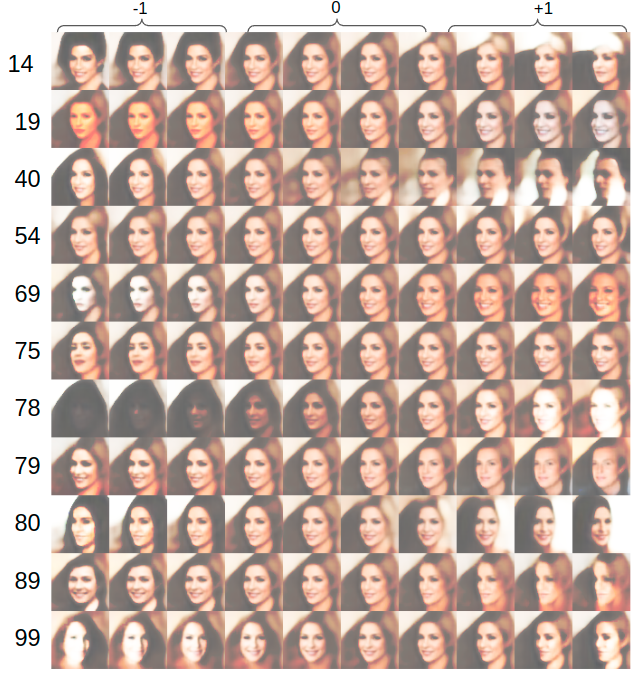}
\caption{Traversal of the latents involed in the identifed subgroups}\label{fig:traversal}
\end{figure*}

\subsection{Understandability of the extracted subgroups }
In order to evaluate if the extracted subgroups are understandable, we examine the 10 subgroups returned by beam search, and we visualize the latents involved in each subgroup. We manipulated the images in the latent space, one latent at a time, and decoded the manipulated images to visualize the meaning of each latent. For this section, due to page limit constraints, we focus only on the fine-tuned model trained with respect to the attractiveness target variable. Visualization for the rest of experiments is provided as supplementary material. Table \ref{beam_content} shows the subgroups along with their quality measures. The first row of the table is an empty subgroup that gives the statistics on the whole population. The traversal for all latents involved in the discovered subgroup is shown in Figure \ref{fig:traversal}. The numbers on the left indicate the corresponding latent identifier. 

The top subgroup, $40==-1.0 \wedge 78==1.0$, means that if latent 40 has a low value and if latent 78 has a high value, then there is 85\% chance that the person is perceived as attractive. This is a very interesting subgroup because on the whole population, as shown in the first row, for a random sample, there is only a 51\% chance of being perceived as attractive. Furthermore, with more than 5000 samples in this subgroup, it has relatively good coverage. By referring to Figure \ref{fig:traversal}, we may observe that latent 40 captures the length of hair, and latent 78 encodes skin color; therefore, considering the assigned value to these concepts, this subgroup indicates that if a person has long hair with a lighter skin color, then they have a higher chance of being perceived as attractive. Similarly, we can see that the second subgroup, $40==-1.0 \wedge 79==-1.0$ , implies that people with longer hair and narrower jaw are more likely to be perceived as attractive. We should note that concept 79, in addition to jaw width, has encoded eye size as well, to some degree. 

\begin{figure*}[h!]
\centering
\includegraphics[scale=0.3]{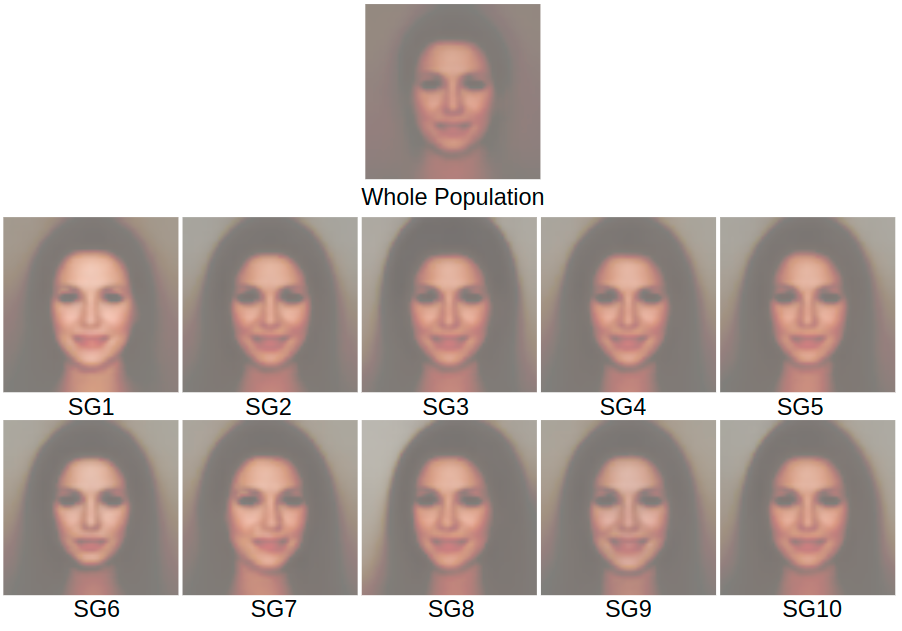}
\caption{Visualization of a typical face in each subgroup and the whole population}\label{fig:avg_subgroups}
\end{figure*}

In order to better understand what features each subgroup is capturing, we have also visualised the average latents, i.e., a typical face, in each subgroup. Figure \ref{fig:avg_subgroups} shows the average latents over the whole population along with average latents for each subgroup. For example, we can see that in subgroup 3, in addition to the long hair which is common among all subgroups (latent 40), the face is narrower in comparison to the typical face in the population.

In conclusion, our analysis has provided anecdotical evidence that the extracted subgroups are human-understandable. We believe that the discovered subgroups can give valuable insights into how the data was labeled and potentially reveal biases of the annotators.

\begin{table}[]
\begin{tabular}{ll|ccccc}
     & Subgroup              & \multicolumn{1}{l}{Coverage} & \multicolumn{1}{l}{Target share} & \multicolumn{1}{l}{\#Samples} & \multicolumn{1}{l}{\#Positive samples} &  \\
  \specialrule{.2em}{0em}{.2em}     
     & Empty                             &  1.000           & 0.513                            & 202599                      & 103833                                 &  \\
1    & 40==-1.0 $\wedge$ 78==1.0  & 0.027                        & 0.859                            & 5428                        & 4663                                   &  \\
2    & 40==-1.0 $\wedge$ 79==-1.0 & 0.025                        & 0.846                            & 5060                        & 4280                                   &  \\
3    & 14==-1.0 $\wedge$ 40==-1.0 & 0.036                        & 0.836                            & 7260                        & 6072                                   &  \\
4    & 40==-1.0 $\wedge$ 75==1.0  & 0.023                        & 0.834                            & 4692                        & 3912                                   &  \\
5    & 40==-1.0 $\wedge$ 89==1.0  & 0.026                        & 0.831                            & 5362                        & 4458                                   &  \\
6    & 40==-1.0 $\wedge$ 69==-1.0 & 0.032                        & 0.827                            & 6568                        & 5431                                   &  \\
7    & 40==-1.0 $\wedge$ 99==-1.0 & 0.026                        & 0.827                            & 5276                        & 4361                                   &  \\
8    & 40==-1.0 $\wedge$ 80==-1.0 & 0.026                        & 0.826                            & 5344                        & 4414                                   &  \\
9    & 19==1.0 $\wedge$ 40==-1.0  & 0.027                        & 0.824                            & 5438                        & 4480                                   &  \\
10   & 40==-1.0 $\wedge$ 54==-1.0 & 0.026                        & 0.815                            & 5327                        & 4344                                   &  \\
Mean &                       & 0.028                        & 0.833                            & 5575.5                      & 4641.5                                 & 
\end{tabular}
\caption{List of identifed subgroups using the fine-tuned model with respect to attractiveness target variable} \label{beam_content}
\end{table}


\subsection{Predictive power of the learnt representation}
As a quantitative way to evaluate the adequacy of the latent representations learned by the subgroup-aware VAE, we trained a classifier on the latent representation for each of the three models. The evaluation was done using a random forest classifier and 5-fold cross-validation. Table \ref{tab:pred_metrics} presents the average values for precision, recall, accuracy, and F1-score.

In general, using the SD loss significantly improves the predictive performance. In terms of precision, fine-tuning the model with subgroup discovery loss either had no impact or improved the performance. With respect to recall, fine-tuning significantly improved the performance, especially for the gender target variable, increasing recall from 0.07 to 0.48. As indicated by the F1 score, the fine-tuned model has a better performance overall in comparison to the baseline. Training the model with SD loss from scratch further improves the performance in all metrics.

We note that the basic VAE model is trained in an unsupervised manner while
the other two VAE models take advantage of the labels, even though they are not using
them in a prediction loss. However, we believe that the presented experiments
still highlight how the representations have changed among the three models. Specifically, these experiments show that adding the subgroup discovery loss results in learning concepts that better explain the variation in the target variable. This in turn, as shown in section \ref{sgd_quality_section}, led to subgroups with higher
quality.


\begin{table}[]
\begin{tabular}{l|llll|llll}
                                & \multicolumn{4}{c}{Attractiveness}           & \multicolumn{4}{c}{Gender}             \\
Method                               & Precision & Recall & Accuracy & F1 & Precision & Recall & Accuracy & F1 \\
\specialrule{.2em}{0em}{.2em}
VAE                                   &      0.63       &    0.8     & 0.66     &  0.70   &     0.93      &     0.07   &   0.61       & 0.13    \\
VAE+SD F.T. &      0.63       &    0.84    & 0.66     &  0.72   &     0.98      &     0.48   &      0.77    & 0.64    \\
VA+SD  &      0.64       &    0.94    & 0.70     &  0.77   &      0.96     &  0.80      &    0.90      &    0.88 \\
\end{tabular}
\caption{The values for precision, recall, accuracy, and F1-score for the predictors trained on features obtained by the three models.} \label{tab:pred_metrics}
\end{table}

\section{Conclusion}
In this paper, we introduced the problem of subgroup discovery in unstructured data and proposed a framework based on variational autoencoders to extract concept-based subgroups. Specifically, we proposed a subgroup-aware VAE, which leverages subgroup discovery in the latent space while training to guide the backpropagation on the path to reaching a subgroup-friendly representation of the data. Experimental results on the CelebA dataset showed that introducing subgroup discovery loss improved the quality of the discovered subgroups significantly in terms of coverage and target share while keeping the reconstruction error at an acceptable level. Furthermore, visualizing the latents confirmed the interpretability of the identified subgroups.

We believe that the proposed method has the potential to be used in many practical domains. The medical domain specifically possesses a wide variety of unstructured data sources and is a prominent example of where our method may be used to extract insights into complex diseases. For example, our method may be applied to a dataset of brain MRI images to identify what parts of the brain have been affected by Alzheimer's disease. While we have provided anecdotical evidence for the understandability of the discovered subgroups, future work should investigate it more systematically including some user study. We believe that subgroup-aware VAEs will have applications in many domains where interpretability of the machine learning model is crucial. These domains have different data types and requirements, which may suggest domain-specific adaptations of our proposed method.


\bibliographystyle{unsrt}  
\bibliography{ref}

\end{document}